# CBAS: CONTEXT BASED ARABIC STEMMER

Mahmoud El-Defrawy, Yasser El-Sonbaty and Nahla A. Belal

College of Computing and Information Technology, AAST, Alexandria, Egypt

*ABSTRACT*

*Arabic morphology encapsulates many valuable features such as word's root. Arabic roots are being utilized for many tasks; the process of extracting a word's root is referred to as stemming. Stemming is an essential part of most Natural Language Processing tasks, especially for derivative languages such as Arabic. However, stemming is faced with the problem of ambiguity, where two or more roots could be extracted from the same word. On the other hand, distributional semantics is a powerful co-occurrence model. It captures the meaning of a word based on its context. In this paper, a distributional semantics model utilizing Smoothed Pointwise Mutual Information (SPMI) is constructed to investigate its effectiveness on the stemming analysis task. It showed an accuracy of 81.5%, with a at least 9.4% improvement over other stemmers.*

## 1.INTRODUCTION

Natural Languages (NLs) are the communication channels between humans. It allows conveying information, exchanging knowledge, and sharing ideas. For many years scientists studied Natural Languages and developed theories and rules that govern the use of Natural Languages, such as Grammar and Morphology. Natural Language Processing (NLP) is the intersection between linguistics, and Computational Science (CS) [1]. NLP allows utilizing linguistics to use Natural Languages as a way of communication with computational devices[1].

The association curve between linguistics and computational sciences has evolved over time. Machine Translation (MT) was one of the first NLP tasks in the 1950s; it began as translation from Russian to English [2]. The progress of MT was limited due to the complexity of linguistics rules, and low computation power at the time[1]. However, Chomsky's theory[3] of natural language's grammar formed the basis for the formation of Backus-Naur Form (BNF). BNF[4] notations are commonly used to represent Context Free Grammar (CFG). CFGs are used to systematically describe, and validate artificial languages, such as programming languages. Using CFGs to describe some aspects of Naturals Languages requires a non-trivial set of rules, which results in some ambiguity due to the unexpected rules' interactions.

The introduction of statistical methods gave some insights for reducing NLP ambiguity[1]. For example, the Probabilistic CFGs extends the traditional CGFs by deducing linguistic rules and assigning weights[5]. Rules and weights are statistically deduced from large annotated corpus. The noticeable improvement in MT sparks the research in NLP [1].

Table 1. Context Matrix Sample.

|  | عجائب (Wonders) | دول (Countries) | جامعة (University) | القاهرة (Cairo) |
|---|---|---|---|---|
| نظم (Systems) | 0 | 5 | 10 | 3 |
| سائح (Tourist) | 8 | 5 | 0 | 14 |
| حكم (Judgement) | 0 | 12 | 0 | 9 |

NLP tends to work with large sets of data from various languages. It raises the need for defining a concise representation for the data while preserving as many of its features as possible. Concise representation is required mostly for any NLP task (word, sentence, or document levels). Stemming is a primary NLP task, and it contributes in many other NLP tasks [6]. Stemming is reducing a word to its basic form[7], while preserving its main characteristics. Many languages define linguistic rules for stemming but not with the same degree[8].

Derivative languages are highly systematic, and highly supportive for stemming analysis. Most of the derivative languages share the property that complex forms are derived from basic ones. Arabic is one of the derivative languages that linguistically supports stemming. The Arabic language is a widely used language[9] and it exists in different formats. For example, Arabic words can be given in the format of separate text or it could be extracted from images[10]. Arabic language defines an accurate set of rules known as morphological rules, or morphology. Morphology accurately describes the formulation of an Arabic word from its basic form. The basic forms are commonly called roots. However, stemming is faced with ambiguity, as most of the NLP tasks.

Various techniques were used to resolve ambiguity. Among which is semantic analysis, that is to capture the intended meaning of a word [11]. Semantic analysis is a very powerful tool to tackle the ambiguity problem but it is very challenging to model. Distributional Semantic (DS)[11] is a type of semantic analysis based on co-occurrence analysis. It represents a word's meaning by its context (surrounding words) distribution as shown in Table 1. For example, the first row of Table 1 shows that the words (Wonders"means ) عجائب and (means "Countries") دول appeared in the context of the word (means "Systems") نظم with frequencies 0 and 5, respectively. Different measures can be computed such as Pointwise Mutual Information (PMI), Positive PMI (PPMI), and Smoothed PMI (SPMI). They measure the correlation between a word, and its context[12,13] .

This paper introduces a context based Arabic stemmer for extracting a word's root. The proposed stemmer (CBAS) explores all possible roots then selects the appropriate root using Distributional Semantics (DS). The DS utilization impact is viewed as a series of comparisons with other stemmers using a manually annotated set of articles. The paper is organized as follows; section 2 is an introduction to Arabic morphology. Section 3 explores related work, and techniques used for constructing stemmers. The description of the proposed stemmer (CBAS) is introduced in section 4. In section 5, a detailed analysis, and evaluation of the proposed stemmer (CBAS) is presented. Finally, a conclusion is presented in section 6.

## 2. BACKGROUND

Regardless of the approach used for developing morphological analyzers, a basic understanding of morphological rules is needed to set expectations, evaluate the results, and design improvements. This section introduces Arabic morphology, and common challenges.

Morphology is the study of word formulation. Arabic morphology is based on the derivation principle, whereas words are acquired from roots. Roots are usually three, four, or five characters.

Roots are the seeds for Arabic words generation. A new word is acquired by modifying its root. For example the word كاتب (kātb, means "Writer") is derived from the root ك ت ب (kāf tāʾ bāʾ , means "Wrote") by adding ا (ʾlf) in the middle.

Not every addition is considered to be valid. Arabic language introduces a set of templates to define valid combinations and additions. Templates are referred to as patterns. It is an ordered sequence of letters. Since patterns work with all roots, a set of letters generically represent roots letters and its order while augmented letters are represented by themselves in correct positions. For example the pattern فاعل (fāʿl, means "Actor") was used to derive the previous word كاتب (kātb, means"Writer") by substituting ف (fāʾ) with ك (kāf), ع (ʿyn) with ت (tāʿ), and ل (lām) with ب (bāʾ),respectively. As noted, roots are commonly written as separated characters to indicate possible insertions[14].

Augmented letters are reflected on the pattern, and finally on the word itself. However, an augmented unit (one, or more augmented letters) which is added in the front, or at the end of a word, is called a prefix, or a suffix addition. In most cases, prefixes and suffixes are not part of a word's meaning, rather they are additional features[14]. For example الكاتب (ʾlkātb, means"The writer") by adding ال (ʾl, means"The") in front of the word. This point of view would substantially reduce the number of enumerated patterns.

As described above, the root-pattern system is simple, elegant, and straightforward. However, the system is faced with morphological challenges, namely vocalization, mutation, and the absence of diacritics (annotation above, or below a word's letter that captures morphological, and grammatical additional features). For example, a letter can change its form due to grammatical or phonological rules. Another challenge of the root pattern system is stopwords, such as connection words that do not obey derivational rules.

The process of deriving back a word to its root (stemming) looks like a straightforward operation, by simply aligning a word to its pattern, and collecting the letters corresponding to ف (fāʾ), ع (ʿyn), and ل (lām) letters. However, due to the challenges described above, a word may be derived back to multiple roots.

## 3. RELATED WORK

Information Retrieval (IR) is one of the early tasks that utilized stemming analysis[7]. But, Stemming analysis is not limited to IR. Stemming analysis has improved many tasks, such as Machine Translation (MT)[15], Sentiment Analysis [16], and many more tasks. This section views common stemming analysis algorithms for different Natural Languages (NLs).

The nature of a language has a great impact on the development of related stemming algorithms. For example, the nature of the English language makes English stemmers concerned with removing word's suffixes only, while removing prefixes may imply a different meaning, such as sufficient and insufficient. Various stemmers were developed for English[8,17]. However, the underlying nature of the language limited its extension for other languages, among which is the Arabic[18] and Urdu[19] languages. However, stemming is not effective in the same degree for all languages[20,21]. Arabic is a morphological rich language which has enriched the Natural Language Processing (NLP)[22-24].
Arabic roots have rich linguistic features; they are semantically representative, derivable, and finite in numbers. Stemmers were developed over the years to take advantages of such features. This section introduces common Arabic stemmers, and their root extraction process which encapsulates semantic decisions. Finally, it introduces semantic analysis techniques independently from stemming analysis.

Khoja stemmer[18] is one of the early and most powerful approaches developed for Arabic stemming[7]. Khoja simulates the linguistic process as much as possible. It removes prefixes, and suffixes from a word often after the normalization process, then matches the resulting word to a pattern, and finally extracts the root. The extracted root gets validated against a list of correct Arabic roots to ensure linguistic correctness. Khoja stemmer[18] resolves ambiguity by defining a set of linguistic paths, or decisions based on various features such as words first character, prefixes, or suffixes length. Additionally, decisions are implicitly ordered whereas the result would be the first correct root. For example, Khoja stemmer handles roots with duplicate letters first.

Root extraction is a highly complex process due to the existence of overlapping rules, which requires more information. A new type of stemming analysis introduced is light stemming. Light stemming is another way of acquiring reduced representation of Arabic words. Light stemming is not as complex as root extraction. It removes prefixes and suffixes only from a word. For example, the word الكاتبون (ʾlkātbūn, means "The writers") would be stemmed to the word كاتب (kātb, means "Writer") instead of ك ت ب (kāf tāʾ bāʾ, means "Writing"). Light stemming is widely used for Information Retrieval (IR)[25]. Light stemming has shown competitive results in IR against root extraction based stemmers [6 ,26]. Light stemmers are relatively faster and efficient, which preserves more specific features of the word, for example, كاتب (kātb, means "Writer") is more related to the word الكاتبون (ʾlkātbūn, means "The writers) than ك ت ب (kāf tāʾ bāʾ, means "Writing"). But, the number of words in Arabic without prefixes and suffixes is far more than the roots listed in Arabic dictionaries[14]. However, there is no explicit evidence that lightly stemmed words are more efficient than roots[26].

ISRI [27] is another linguistic based Arabic stemmer that roughly uses the same sequence used by Khojas stemmer [18]. But, the main difference is that ISRI does not linguistically validate the extracted root, no dictionary is used, and organizes the defined pattern set as sub-groups whereas each sub-group has common features. Besides, changes to the normalization process, prefixes, and suffixes handling. The main goal of ISRI is to get the minimum representation of a given word. Due to various changes, and prioritization, ISRI resolves ambiguity differently, by changing the order of applying morphological rules. But, ISRI still makes static decisions. Most likely, ISRI would be used for information retrieval rather than linguistic based tasks.

Tashaphyne [28] is another Arabic stemmer. It mainly supports light stemming (removing prefixes and suffixes). It follows the same approach used by Khoja [18], and ISRI [27] stemmers. It can be used for root extraction as well.

Darwish[29] utilizes the existing word-root pairs used to construct the Finite State Transducer (FST) like stemmers. It uses a learning based technique to not only infer Arabic patterns, but also to rank extracted roots. This methodology enumerates possible roots like FSTs [30] approaches, but additionally gives a preference to the extracted roots. It is another way to handle ambiguity other than static rules. However, part of the inferred patterns would be inadequate due to cases such as vocalization, and mutation. And, the ranking of roots is based on inferred patterns frequency, neglecting the words features. Later, this approach has been modified to handle light stemming (prefixes and suffixes removal) [7].

Arabic grammar has high influence on morphological analysis[14]. ElixirFM [31] employs syntactic features to enhance morphological results. It takes advantage from Prague Arabic Dependency Treebank (PADT)[32] to acquire syntactic features and other morphological features from BuckWalter [33] stem dictionary. ElixirFM[31] defines a set of morphological rules to extract possible stems while ranking is used to disambiguate the extracted roots, or stems using the underlying data.

MADAMIRA[15] morphological analyzer consists of two tools MADA [34], and AMIRA [35]. MADAMIRA [15] takes the advantage of large annotated corpus by using machine learning techniques such as Support Vector Machine[35]. It combines several tools such as word segmentation, Part of Speech Tagging (POST), and light stemming. It is different from the previous approaches, since it does not explicitly define morphological rules.

Stemmers are part of many NLP tasks. For example, in sentiment analysis they employ the use of classifiers such as Naive Bayes or Support Vector Machines (SVMs) to perform classification on sets of test samples given a tagged training set and rich feature sets for various tasks[16,36,37], question and answer systems [38,39], and many more. However, stemming is not limited to NLP. It has been used for Information Retrieval (IR), and most of the stemmers are evaluated indirectly using IR benchmarks [7]. Many IR experiments [6,26,27] showed that Arabic roots had improved the Arabic IR.

## 4. PROPOSED STEMMER (CBAS)

The proposed stemmer, Context-Based Arabic Stemmer (CBAS), utilizes distributional similarity of a word's context to gain additional information about its semantic. This information assists with the selection of the correct root by excluding semantically irrelevant candidate roots within the context.

This section introduces the main phases of the proposed stemmer (CBAS), context matrix construction, roots generation, and root selection, whereas each phase consists of a set of steps. The proposed algorithm is shown in Fig. 1 and Fig. 2

```
1: procedure CONTEXT MATRIX CONSTRUCTION(documents)
2:     normalize all words.
3:     define context window of size n words.
4:     for all documents do
5:         slide the winodw over the document.
6:         remove stopwords
7:         update associated counts of word at position i and rest words in the window.
8:     end for
9: end procedure
```

Fig.1: Context Matrix Construction Algorithm

```
1:  procedure STEM(word, context)
2:      remove stopwords
3:      normalize the given word, and its context.
4:      segment word into prefix, infix, and suffix
5:      validate segments using prefixes, and suffixes lists.
6:      for all patterns do
7:          match infix to pattern with the same length.
8:          extract primary root.
9:          work out weak letters if exists.
10:         validate roots against dictionary.
11:     end for
12:     for all correct roots do
13:         generate set of words
14:         compute the average spmi with its context
15:     end for
16:     select the root with heighest average spmi
17: end procedure
```

Fig.2: Stemmer's Algorithm

### 4.1. Data Resources

Predefined linguistic data is an essential part of the proposed stemmer (CBAS). This section introduces the data defined by CBAS. The Arabic word consists of three parts prefix, infix, and suffix[14]. Prefixes and suffixes are a set of features that can be added to a word, such as the definite article ال ('l, means"The") or connected pronouns هم (hm, means "Them"). Prefixes and suffixes lists contain individual and compound letters that could appear in the front or the end of the Arabic word. There is also a list of Arabic patterns which is used to extract possible Arabic roots. The final list is the Roots' dictionary which has been extracted from the Khoja stemmer [18] to validate the extracted root. The prefixes, suffixes, patterns, and dictionary lists are being used for the roots' generation phase.

Previous lists are commonly defined for Arabic stemmers. However, CBAS uses a raw data set which consists of a set of Articles that have been extracted from Omani newspapers[40]. The dataset contains 20291 articles from various topics, for example, culture and sport[40], which represents a wide range of the Arabic language current usage. The raw dataset plays a central role in selecting semantically correct roots, which differs from other used stemmers which do not commonly employ context in their algorithms.

### 4.2. Context Matrix Construction

Context Matrix is a powerful and flexible tool to acquire some semantic properties [11]. It defines a window of $n$ words, where the target word is at position $i$, and the rest of the surrounding words are its context [11]. The window slides over the corpus associating the target word with its context distribution as show in Table 1. Various measures can be computed from the context matrix, and employed in different tasks.

### 4.3. Root Generation

It is automation for root extraction. However, unlike the manual process, it extracts all possible roots. It consists of three major sub processes, word segmentation, pattern matching, and root validation.
– Word segmentation breaks a word into all possible three parts, prefix, suffix, and infix, using the predefined prefixes, and suffixes lists.
– Pattern matching matches the infix obtained in word segmentation with one or more patterns with respect to its length. For each matched pattern, roots characters are collected, and passed to dictionary validation. It also handles weak letters, stopwords, and some other linguistic cases.
– Dictionary validation ensures that the extracted roots are linguistically correct. It validates extracted roots against a list of correct Arabic roots.

Dictionary is not sufficient to generate only one correct root, due to various linguistic cases, roots' generation has the potential of extracting one, or more correct roots.

### 4.4. Roots' Selection

This section will utilize the context matrix to select an appropriate root from two or more candidate roots. Pointwise Mutual Information (PMI) [12,13] measures the correlation between two or more words. Since some words can produce two, or more root candidates. The proposed algorithm (CBAS) uses a variation of PMI, Smoothed PMI (SPMI) [12] to handle sparse matrices. As shown in Table 2, SPMI achieved the highest accuracy of 81.5% when compared to PMI and PPMI, where the achieved accuracy was 78.84% and 79.49%, respectively. This is due to the fact that SPMI overcomes the tendency towards rare co-occurrence events, which is a side effect of PMI [12].

SPMI is utilized to measure the correlation between the generated roots, and its previous context. To take advantage of the underlying matrix, set words are derived for each candidate root, in addition to the root itself. For each derived word, the average SPMI is computed with the previous word (as context). The root with the highest average correlation to its context is then selected.

## 5. RESULTS AND EVALUATION

IR is the common methodology for evaluating a new stemmer because of the lack of stemmed benchmarks[21]. This section introduces the validation dataset, evaluation measures, and finally the experimental results.

### 5.1. Validation Dataset

Direct evaluation is important to show the stemming accuracy, and potential improvements. A manually annotated dataset has been provided to measure the stemmer accuracy, and compare with other stemmers. The dataset is part of the Intentional Corpus of Arabic (ICA) [41].Various Arabic resources have contributed in collecting the ICA such as newspapers, books, and magazines. It has been constructed to provide an appropriate representation to Arabic language in Modern Standard Arabic (MSA) [41]. The dataset consists of 10302 tokens associated with various features. There exist 3629 unique word-root pairs, while other words do not have roots associated due to the existence of stopwords, and non-Arabic words. The dataset contains 8941 words after stopwords removal. This is shown in Fig 3.

International Journal on Natural Language Computing (IJNLC) Vol. 4, No.3,June 2015| Doc_ID | Word | Lemmaid | Pr1 | Pr2 | Tags | Suf1 | Suf2 | Root |
|---|---|---|---|---|---|---|---|---|
| 214 | الجمعة | jumoEap | Al/DET | | NOUN_PROP | | | jmE |
| 215 | 24 | | | | Num | | | |
| 216 | نوفمبر | nuwfamobir | | | NOUN_PROP | | | |
| 217 | بالمسرح | masoraH | bi/PREP | Al/DET | NOUN | | | srH |
| 218 | الصغير | Sagiyr | Al/DET | | ADJ | | | Sgr |
| 219 | لدار | dAr | li/PREP | | NOUN | | | dwr |
| 220 | الأوبرا | >uwbirA | Al/DET | | NOUN | | | |
| 221 | ، | | | | Punc | | | |
| 222 | وقال | qAl-u | wa/CONJ | | PV | a/PVSUFF_SUE | | qwl |
| 223 | د | | | | ABBREV | | | |
| 224 | . | | | | Punc | | | |
| 225 | مرعي | maroEiy | | | NOUN_PROP | | | rEy |
| 226 | مدكور | madokuwr | | | NOUN_PROP | | | |
| 227 | عضو | EuDow | | | NOUN | | | EDw |
| 228 | الاتحاد | {it~iHAd | Al/DET | | NOUN | | | wHd/'Hd |
| 229 | المصري | miSoriy~ | Al/DET | | ADJ | | | mSr |
| 230 | : | | | | Punc | | | |
| 231 | إن | <in~a | | | SUB_CONJ | | | |
| 232 | افتتاح | {ifotitAH | | | NOUN | | | ftH |
| 233 | الدورة | daworap | Al/DET | | NOUN | ap/NSUFF | | dwr |
| 234 | الـ 23 | | | | Num | | | |
| 235 | ... | mu&tamar | li/PREP | Al/DET | NOUN | | | |

Fig.3: Validation Dataset

## 5.2. Evaluation Criteria

Stemming is beneficial for many tasks, where every task uses the roots in a different way. For example, IR uses roots as a cluster representative to group related words, while sentiment analysis is more concerned with the linguistic accuracy of a root. A set of metrics were used to measure different usages, and compare them with other stemmers.

Stemming accuracy is one of the basic measures for the effectiveness of the stemmer. It is defined as the ratio between the number of correctly stemmed words, and the number of the words in the complete dataset.

Collecting related words under the same group is important for tasks such as IR. There are two variations of grouping related words. First, words can be grouped correctly under a semantically correct root; this is referred to it as classification. While the second is to group related words together, not necessarily under a correct root, and this is referred to as clustering. Standard metrics for classification, and clustering are: *accuracy, precision, recall, and $F_1$ measure*, and are defined as follows [42, 43]:

$$accuracy = \frac{1}{n} \sum_{i=1}^{n} \frac{|X_i \cap Y_i|}{|X_i \cup Y_i|}$$

$$precision = \frac{1}{n} \sum_{i=1}^{n} \frac{|X_i \cap Y_i|}{|Y_i|}$$

$$recall = \frac{1}{n}\sum_{i=1}^{n}\frac{|X_i \cap Y_i|}{|X_i|}$$

$$F_1 measure = \frac{1}{n}\sum_{i=1}^{n}\frac{|X_i \cap Y_i|}{|X_i|+|Y_i|}$$

Where
- $n$ is the number of extracted roots.
- $X$ is the set of extracted root.
- $X_i$ is an individual extracted root.

And
- $Y$ is the set of extracted root.
- $Y_i$ is an individual valid root.

## 5.3. Results

The complete 8941 words were used to test the proposed stemmer (CBAS), with a window size $n=3$, then the set was reduced to a set of unique word-root pairs to be compared with other stemmers. Table 2 shows the comparison between the proposed stemmer (CBAS) and other stemmers. It shows that the proposed stemmer (CBAS) achieved an accuracy of 81.5% with an improvement of 9.4%, 67.3%, and 51.2% over Khoja, ISRI, and Tashphanye stemmers, respectively. Accuracy enhancement is due to exploring various possibilities of roots. Such exploration would not be possible without distributional semantics, which provides a dynamic and robust way for selecting an appropriate root.

Table 3 and Table 4; show the performance of the proposed stemmer (CBAS) when using it as a grouping mechanism. Table 3 clarifies that the proposed stemmer (CBAS) has a higher potential to linguistically group Arabic words than other stemmers. CBAS outperformed other stemmers in the classification task, with an accuracy of 65.45%. While Table 4 shows that the proposed stemmer (CBAS) has potential improvements in non-linguistic based tasks, achieving an accuracy of 73.83% in clustering.

By comparing linguistic (classification), and non-linguistic (clustering) grouping measures, there is an increase in all corresponding measures. This is due to that some clusters were correctly formulated irrespective to the clusters seeds. Classification and clustering measures show the superiority of the CBAS over other stemmers. This indicates the beneficial features of the CBAS for the IR task.

Table 2. Stemmers Linguistic Accuracy

| Stemmer | Linguistic Accuracy |
|---|---|
| Khoja | 72.1% |
| ISRI | 14.2% |
| Tashaphanye | 30.3% |
| CBAS-PMI | 78.84% |
| CBAS-PPMI | 79.49% |
| CBAS | 81.5% |

Table 3.  Stemmers Classification Measures

| Stemmer | Accuracy | Precision | Recall | $F_1$ measure |
|---|---|---|---|---|
| Khoja | 57.53% | 57.53% | 59.59% | 58.55% |
| ISRI | 10.43% | 10.43% | 10.49% | 10.46% |
| Tashaphanye | 25.07% | 25.07% | 25.15% | 25.11% |
| CBAS | 65.45% | 65.45% | 68.23% | 66.51% |

Table 4.  Stemmers Clustering Measures

| Stemmer | Accuracy | Precision | Recall | $F_1$ measure |
|---|---|---|---|---|
| Khoja | 71.71% | 93.09% | 75.74% | 83.52% |
| ISRI | 12.59% | 69.40% | 13.34% | 22.27% |
| Tashaphanye | 32.25% | 72.54% | 37.03% | 49.03% |
| CBAS | 73.83% | 93.71% | 75.46% | 84.50% |

## 6. CONCLUSION

Many stemmers were developed to gain the rich linguistic features provided by the roots. Most of the stemmers made explicit decisions, statistical-based or linguistic-based, to select only one root. Other stemmers used ranking to express their selection preference rather than selecting a single root. However, at the very end, a single root would be chosen. Static decisions are very appropriate for common and frequent cases. However, adding other features such as syntactic and manual annotations would also be valuable. The introduced stemmer employs distributional similarity to handle incorrect roots selection, which is a side effect of root generation phase. The existence of robust filtering mechanisms, such distributional analysis, allows exploring various roots. Distributional analysis has several advantages. It can be computed for any corpus and any language, and it is relatively fast and inexpensive to construct compared to manually annotated corpus. Distributional semantics covers many relations between words, and it is robust against any preferences, or missing information. It is also very adaptive to context changes, which makes it suitable for many topics. However, distributional analysis is not as accurate as manually annotated data; hence, the word generation process was added to the roots selection phase to tolerate possible errors. The previous techniques were compared to the proposed stemmer (CBAS) results. CBAS shows an accuracy of 81.5% with an improvement of 9.4%, 67.3%, and 51.2% over Khoja, ISRI, and Tashphanye stemmers, respectively. CBAS also shows an improvement in classification and clustering, with an accuracy of 65.45% and 73.83%, respectively. Results indicate that the proposed stemmer (CBAS) enhances stemming and other related tasks. CBAS represents a methodology for capturing a word's context and makes decisions based on it. CBAS could change its behaviour based on the underlying data which could be specialized in a sub domain of the Arabic language. The statistical model used by CBAS is relatively simple. It incorporates important information (context) of a word which would be a complex process to include in a rule based stemmer. The statistical model reduces linguistic complexity of representing various linguistic cases. It also prevents unexpected interactions and prioritization schemes for ordering the rules.

## REFERENCES

[1] P. M. Nadkarni, L. Ohno-Machado, and W. W. Chapman, "Natural language processing: an introduction," Journal of the American Medical Informatics Association, vol. 18, pp. 544-551, 2011.
[2] J. Hutchins, "The first public demonstration of machine translation: the Georgetown-IBM system, 7th January 1954," noviembre de, 2005.
[3] N. Chomsky, "Three models for the description of language," Information Theory, IRE Transactions on, vol. 2, pp. 113-124, 1956.
[4] A. Aho, "R. Sethi, and J. D. Ullman," Compilers: Principles, Techniques, and Tools, 1988.
[5] D. Klein and C. D. Manning, "Accurate unlexicalized parsing," in Proceedings of the 41st Annual Meeting on Association for Computational Linguistics-Volume 1, 2003, pp. 423-430.
[6] M. Aljlayl and O. Frieder, "On Arabic search: improving the retrieval effectiveness via a light stemming approach," in Proceedings of the eleventh international conference on Information and knowledge management, 2002, pp. 340-347.
[7] I. A. Al-Sughaiyer and I. A. Al-Kharashi, "Arabic morphological analysis techniques: A comprehensive survey," Journal of the American Society for Information Science and Technology, vol. 55, pp. 189-213, 2004.
[8] M. F. Porter, "Snowball: A language for stemming algorithms," ed, 2001.
[9] J. Xu, A. Fraser, and R. Weischedel, "Empirical studies in strategies for Arabic retrieval," in Proceedings of the 25th annual international ACM SIGIR conference on Research and development in information retrieval, 2002, pp. 269-274.
[10] R. Fathalla, Y. El Sonbaty, and M. A. Ismail, "Extraction of Arabic Words form Complex Color Images," in 9th IEEE International Conference on Document Analysis and Recognition (ICDAR 2007), Brazil, pp. 1223-1227.
[11] C. Akkaya, J. Wiebe, and R. Mihalcea, "Utilizing semantic composition in distributional semantic models for word sense discrimination and word sense disambiguation," in Semantic Computing (ICSC), 2012 IEEE Sixth International Conference on, 2012, pp. 45-51.
[12] D. Jurafsky. Word Senses and Word Relations.
[13] G. Bouma, "Normalized (pointwise) mutual information in collocation extraction," Proceedings of GSCL, pp. 31-40, 2009.
[14] K. C. Ryding, A reference grammar of modern standard Arabic: Cambridge university press, 2005.
[15] A. Pasha, M. Al-Badrashiny, M. Diab, A. El Kholy, R. Eskander, N. Habash, et al., "Madamira: A fast, comprehensive tool for morphological analysis and disambiguation of arabic," in Proceedings of the Language Resources and Evaluation Conference (LREC), Reykjavik, Iceland, 2014.
[16] S. M. Oraby, Y. El-Sonbaty, and M. A. El-Nasr, "Exploring the Effects of Word Roots for Arabic Sentiment Analysis," in International Joint Conference on Natural Language Processing, Nagoya, Japan, 2013, pp. 471-479.
[17] J. B. Lovins, Development of a stemming algorithm: MIT Information Processing Group, Electronic Systems Laboratory, 1968.
[18] S. Khoja and R. Garside, "Stemming arabic text," Lancaster, UK, Computing Department, Lancaster University, 1999.
[19] M. S. Husain, "An unsupervised approach to develop stemmer," International Journal on Natural Language Computing, vol. 1, pp. 15-23, 2012.
[20] D. Harman, "How effective is suffixing?," JASIS, vol. 42, pp. 7-15, 1991.
[21] I. Smirnov, "Overview of stemming algorithms," Mechanical Translation, vol. 52, 2008.
[22] Y. Benajiba, M. Diab, and P. Rosso, "Arabic named entity recognition using optimized feature sets," in Proceedings of the Conference on Empirical Methods in Natural Language Processing, 2008, pp. 284-293.
[23] K. Darwish and D. W. Oard, "CLIR Experiments at Maryland for TREC-2002: Evidence combination for Arabic-English retrieval," DTIC Document2003.
[24] L. S. Larkey and M. E. Connell, "Arabic information retrieval at UMass in TREC-10," DTIC Document2006.
[25] L. S. Larkey, L. Ballesteros, and M. E. Connell, "Light stemming for Arabic information retrieval," in Arabic computational morphology, ed: Springer, 2007, pp. 221-243.

[26] L. S. Larkey, L. Ballesteros, and M. E. Connell, "Improving stemming for Arabic information retrieval: light stemming and co-occurrence analysis," in Proceedings of the 25th annual international ACM SIGIR conference on Research and development in information retrieval, 2002, pp. 275-282.
[27] K. Taghva, R. Elkhoury, and J. Coombs, "Arabic stemming without a root dictionary," in null, 2005, pp. 152-157.
[28] T. Zerrouki. (2010). Tashaphyne, Arabic light stemmer/segment.
[29] K. Darwish, "Building a shallow Arabic morphological analyzer in one day," in Proceedings of the ACL-02 workshop on Computational approaches to semitic languages, 2002, pp. 1-8.
[30] K. R. Beesley, "Arabic morphological analysis on the Internet," in Proceedings of the 6th International Conference and Exhibition on Multi-lingual Computing, 1998.
[31] O. Smrž, "Elixirfm: implementation of functional arabic morphology," in Proceedings of the 2007 Workshop on Computational Approaches to Semitic Languages: Common Issues and Resources, 2007, pp. 1-8.
[32] O. PetrZemánek, "Prague Arabic Dependency Treebank: A Word on the Million Words."
[33] T. Buckwalter, "Buckwalter {Arabic} Morphological Analyzer Version 1.0," 2002.
[34] N. Habash, O. Rambow, and R. Roth, "MADA+ TOKAN: A toolkit for Arabic tokenization, diacritization, morphological disambiguation, POS tagging, stemming and lemmatization," in Proceedings of the 2nd International Conference on Arabic Language Resources and Tools (MEDAR), Cairo, Egypt, 2009, pp. 102-109.
[35] M. Diab, K. Hacioglu, and D. Jurafsky, "Automated methods for processing arabic text: From tokenization to base phrase chunking," Arabic Computational Morphology: Knowledge-based and Empirical Methods. Kluwer/Springer, 2007.
[36] S. N. Saleh and Y. El-Sonbaty, "A feature selection algorithm with redundancy reduction for text classification," in Computer and information sciences, 2007. iscis 2007. 22nd international symposium on, 2007, pp. 1-6.
[37] S. Oraby, Y. El-Sonbaty, and M. A. El-Nasr, "Finding Opinion Strength Using Rule-Based Parsing for Arabic Sentiment Analysis," in Advances in Soft Computing and Its Applications, ed: Springer, 2013, pp. 509-520.
[38] A. M. Ezzeldin, M. H. Kholief, and Y. El-Sonbaty, "ALQASIM: Arabic language question answer selection in machines," in Information Access Evaluation. Multilinguality, Multimodality, and Visualization, ed: Springer, 2013, pp. 100-103.
[39] A. M. Ezzeldin, Y. El-Sonbaty, and M. H. Kholief, "Exploring the Effects of Root Expansion, Sentence Splitting and Ontology on Arabic Answer Selection," Natural Language Processing and Cognitive Science: Proceedings 2014, p. 273, 2015.
[40] M. Abbas, K. Smaïli, and D. Berkani, "Evaluation of Topic Identification Methods on Arabic Corpora," JDIM, vol. 9, pp. 185-192, 2011.
[41] S. Alansary, M. Nagi, and N. Adly, "Building an International Corpus of Arabic (ICA): progress of compilation stage," in 7th international conference on language engineering, Cairo, Egypt, 2007, pp. 5-6.
[42] S. Godbole and S. Sarawagi, "Discriminative methods for multi-labeled classification," in Advances in Knowledge Discovery and Data Mining, ed: Springer, 2004, pp. 22-30.
[43] M. Hillenmeyer. Machine Learning.